Accepted and presented at the 10th International Conference on Road Safety and Simulation (RSS 2026), Naples, Italy. Proceedings publication forthcoming.

# Virtual Testing of Automated Driving Systems through Credible Simulations

Riccardo Donà[a,*], Espedito Rusciano[b], Biagio Ciuffo[a]

[a]*Joint Research Centre for the European Commission, Ispra (VA) 21027, Italy*
[b]*Joint Research Centre for the European Commission, Petten 1755-ZG, Netherlands*

**Abstract**

Simulation is increasingly used to support safety-related decision-making in road transport, particularly for the assessment and approval of automated driving systems (ADS). The complexity of ADS behavior and size of their operational design domains make exclusive reliance on physical testing impractical, leading to extensive use of virtual testing (VT) during the approval phase. This shift raises critical questions regarding the credibility of modelling and simulation (M&S) results used to support road safety decisions. Current VT accreditation approaches in the ADS domain typically rely on validation-only practices, which have been shown to scale poorly when applied to complex, multi-tool simulation environments. To address this limitation, this paper proposes a risk-based framework for assessing the credibility of simulation toolchains used in ADS safety evaluation, drawing inspiration from established practices in other safety-critical domains, notably NASA's STD-7009 for models and simulations. The framework extends traditional verification and validation (V&V) by explicitly linking credibility requirements to the intended use of simulation outputs and to the safety criticality of the decisions they support within the approval process. It provides a lifecycle-oriented assessment scheme integrating toolchain management, modelling assumptions and limitations, verification, validation, and sensitivity analysis. Credibility acceptance thresholds are defined proportionally, allowing differentiated requirements depending on whether simulation is used for exploratory safety analysis, partial decision support, or as a substitute for physical testing. While demonstrated for ADS, the proposed approach is directly applicable to road safety and simulation studies where VT plays a central role in safety assessment and regulatory decision-making.

---

* Corresponding author. Tel.: 0332 789111 fax:.
*E-mail address:* riccardo.dona@ec.europa.eu

## 1. Introduction

Automated Driving Systems (ADSs) are expected to significantly improve future mobility, particularly in terms of safety and accessibility, but their deployment depends on demonstrating safe operation within the Operational Design Domain (ODD). Unlike conventional vehicles, ADSs pose major challenges due to their complexity, use of AI, and exposure to an almost infinite set of scenarios, making physical testing alone insufficient.

To address this, the United Nations Economic Commission for Europe (UNECE) introduced the multi-pillar New Assessment Test Method (NATM), shown in Fig 1, which covers the entire ADS lifecycle, including organizational processes through a Safety Management System, safety case development, multiple testing approaches, and in-service monitoring. Inspired by practices from aviation, maritime, and nuclear sectors (Donà et al. 2022), NATM combines proven regulatory concepts with industrial feasibility. For testing, NATM allows safety evidence to be generated through virtual, proving-ground, and real-world testing, with virtual testing offering unmatched scalability and flexibility, from Model-in-the-Loop to Hardware-in-the-Loop configurations. However, this flexibility comes at the cost of potential reduced fidelity due to modeling assumptions.

Reduced fidelity is a well-known risk in simulation-based testing, and several industries have addressed it by developing qualification approaches for simulation toolchains. Traditionally, accreditation relies on validation, demonstrating that simulation errors remain below predefined thresholds, an approach also used in automotive regulations. However, validation-based accreditation has clear limitations: error thresholds are often arbitrary, and validation alone is insufficient if models are applied outside their validated domain. History has shown that misuse of validated models can lead to severe consequences, as illustrated by the 2003 Space Shuttle Columbia disaster.

This accident motivated the development of NASA STD-7009 (Babula et al. 2009; NASA 2016), which introduced the concept of Modelling and Simulation (M&S) credibility, defined as the ability to trust simulation results for a specific intended use. The standard promotes a risk-based approach, where validation and verification efforts are proportional to the criticality of decisions supported by simulation, rather than fixed acceptance thresholds. This paradigm strongly influenced other safety-critical sectors, such as aviation, but had limited adoption in automotive.

Recognizing this gap, UNECE integrated M&S credibility into the NATM framework for ADS certification. This paper summarizes how NASA STD-7009 principles were adapted to automotive needs, resulting in the “test environment assessment” pillar of NATM, which aims to ensure virtual testing is as trustworthy as physical testing while preserving flexibility. The remainder of the paper discusses the underlying risk-based approach and future perspectives for virtual testing credibility in automotive regulation.

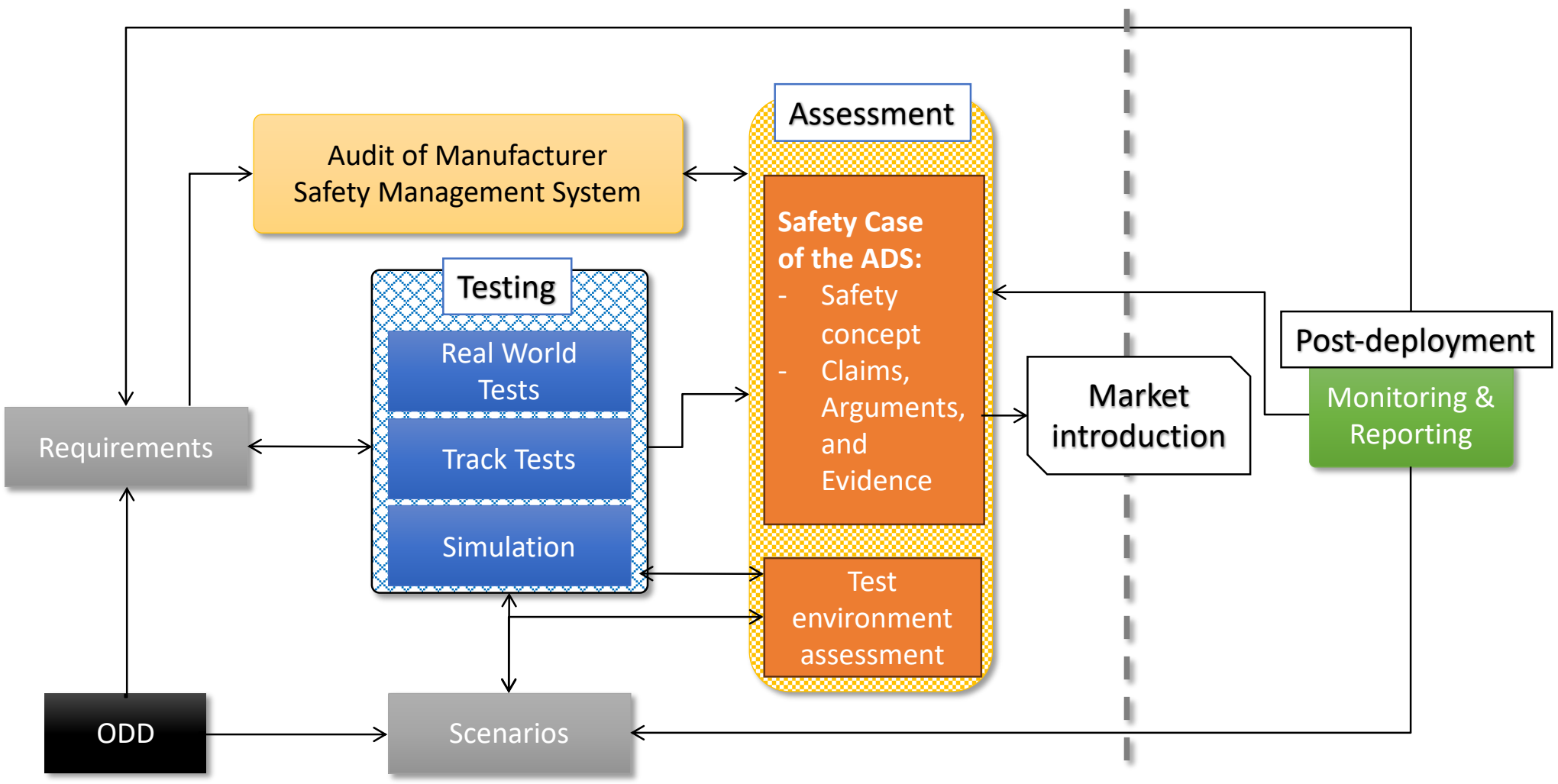

Fig 1. NATM Adapted from (UNECE 2023).

## 2. Methodology

The development of the framework was grounded in an extensive review of the scientific and regulatory literature, complemented by structured consultations with Subject Matter Experts (SMEs) from both academia and industry. This process revealed several challenges in defining regulatory criteria for the qualification of simulation toolchains used in ADS virtual testing. A primary difficulty concerned the need to reconcile the rapid and continuous evolution of M&S toolchains with the inherently slower pace of regulatory and policy cycles. In parallel, stakeholders consistently highlighted the importance of preserving methodological flexibility, so that the framework could remain applicable across a wide spectrum of simulation environments, ranging from Model-in-the-Loop (MiL) and Hardware-in-the-Loop (HiL) to Vehicle-Hardware-in-the-Loop (VeHiL) configurations (Dona et al. 2022; Riedmaier et al., n.d.).

A further challenge stems from the variability in model abstraction levels inherent to virtual testing. Using sensor modeling as an illustrative example, implementations may span from highly abstract object-list representations to physics-based models offering maximum fidelity. The coexistence of such heterogeneous modeling choices within and across simulation toolchains effectively precludes the use of traditional validation approaches based on fixed acceptance thresholds. In this context, a framework inspired by NASA STD-7009 emerged as the most suitable foundation, as it enables the assessment of M&S credibility across different toolchain realizations while enforcing rigorous, use-dependent requirements aligned with regulatory needs.

Building on an agreed high-level methodology, the activities therefore focused on tailoring an automotive-oriented, NATM-compatible M&S credibility framework capable of integration within the broader NATM ADS validation process. Although validation and verification practices are well established in automotive M&S, the prevalent reliance on third-party simulation tools and data introduced additional challenges related to transparency, configuration control, and traceability, all of which were identified as critical enablers of credibility. Furthermore, in line with existing standards, the framework explicitly addresses the documentation of personnel competence to ensure that engineers are aware of the limitations and assumptions embedded in the M&S toolchain. Given the absence of a formal Safety Management System (SMS) requirement in the automotive sector, and the limited coverage of task-specific competencies within traditional SMS frameworks, the explicit qualification of training and expertise was incorporated as an integral element of the proposed methodology.

## 3. Results

This section presents the resulting M&S credibility framework developed for risk-based assessment of simulation toolchains in ADS approval, schematically illustrated in Fig 2. It is structured around four main sub-pillars, which together define a comprehensive approach to simulation toolchain credibility assessment:

- M&S toolchain Management;
- M&S toolchain Analysis;
- M&S toolchain Verification;
- M&S toolchain Validation.

The requirements associated with each sub-pillar are expected to be fulfilled by the manufacturer, with the corresponding evidence systematically documented in a dedicated *credibility handbook*. The handbook will contain the summary of the evidence backing the different sub-pillars. Fig 2 also clarifies the role of the assessor in evaluating compliance with the proposed framework. In particular, the assessor is expected to:

- review the credibility handbook, and
- have access to the integrated M&S toolchain in order to perform independent virtual tests.

The credibility assessment process begins with the explicit definition, by the manufacturer, of the *intended use* of the M&S toolchain within the overall ADS certification strategy. Depending on the ADS use case and certification approach, manufacturers may pursue multiple intended uses for the same simulation toolchain, ranging from exploratory analysis to direct support of approval decisions.

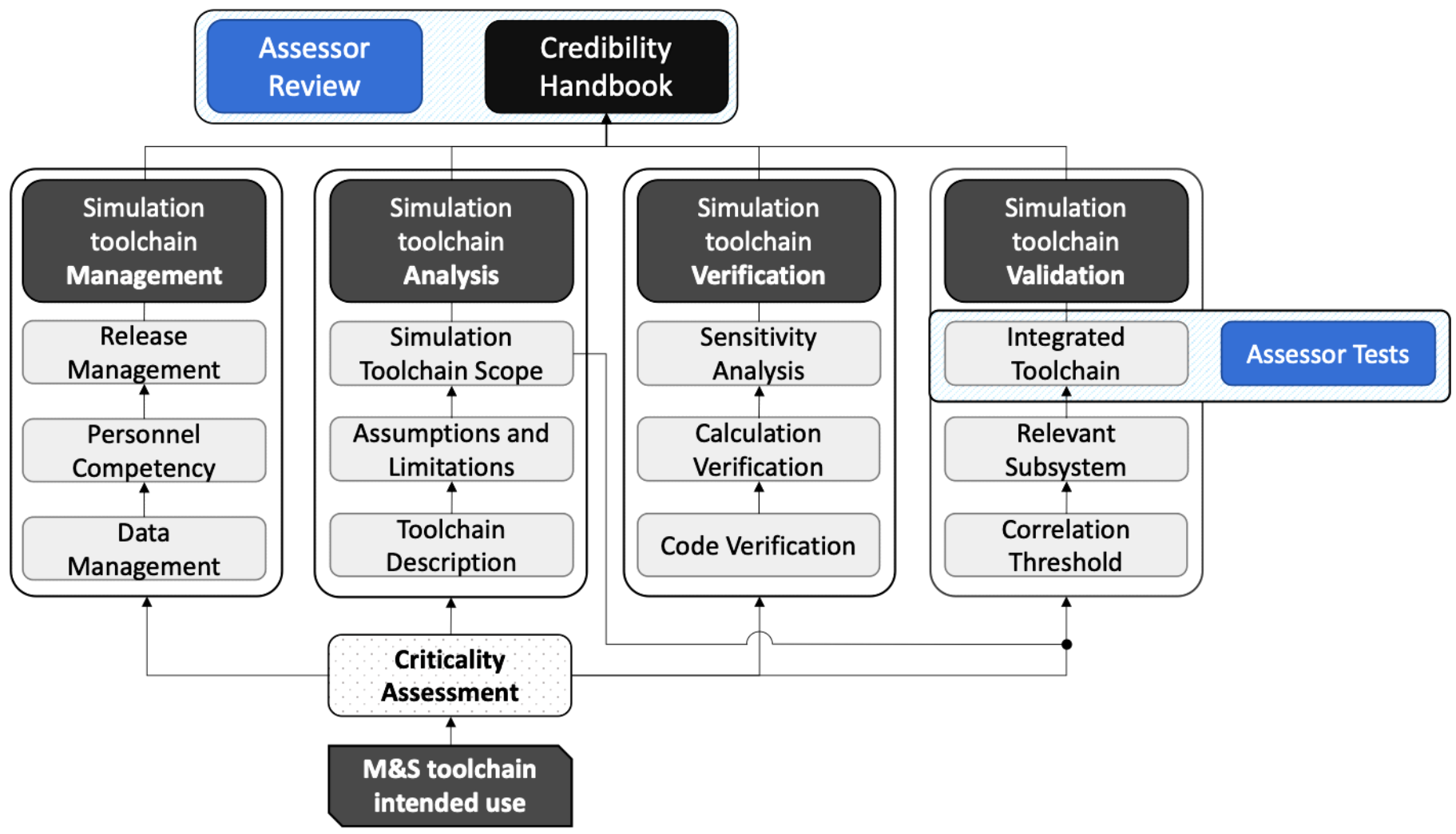

Fig 2. Credibility framework.

Closely linked to the intended use is the definition of the *criticality* of the M&S toolchain, which represents a cornerstone of the proposed risk-based approach. The criticality assessment by the approval authority requires practitioners to explicitly reason about, and document, the implications of modelling assumptions, toolchain configurations, and selected verification and validation techniques. An example of such an assessment is provided in Fig 3.

Consistent with established risk-based practices, the criticality evaluation is based on the relationship between potential consequences and the degree of controllability or exposure. In this context, the *influence on ADS* score represents the extent to which the M&S toolchain affects decision-making within the approval process, while the *decision consequence* reflects the potential impact on human safety, goods, and/or the operational capabilities of the ADS. Based on this assessment, the M&S toolchain is assigned to one of three criticality classes:

- **red**: high criticality, requiring a full credibility assessment in accordance with the proposed guidelines;
- **yellow**: medium criticality, where a full assessment may be required at the discretion of the technical authority;
- **green**: low criticality, for which a full credibility assessment is not required.

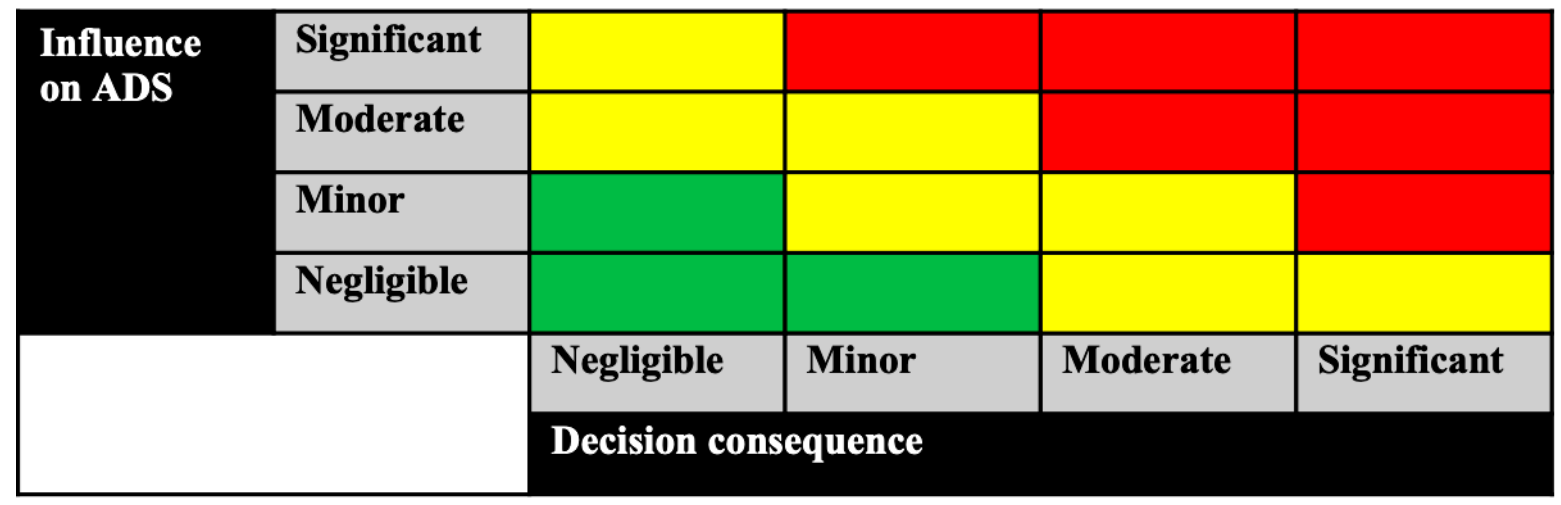

Fig 3. Criticality assessment. Authors adaptation based on (Ciuffo Biagio and Riccardo Donà 2024) .

### *3.1. M&S Management*

The M&S management pillar represents a key structural component of the proposed credibility framework and addresses aspects that are often underrepresented in automotive simulation practice. Rather than focusing solely on validation outcomes, this pillar emphasizes systematic management of the simulation toolchain as a prerequisite for credible safety assessment. Its definition is informed by the results of the criticality assessment, which provide the basis for tailoring management requirements and defining proportional acceptance thresholds. The management pillar is articulated into three sub-classes that collectively contribute to the overall credibility score associated with management-related aspects.

A first element of the M&S management pillar concerns the handling of data used as input to, and generated by, the simulation toolchain. For input data, the framework requires documentation of the following attributes:

- **Data pedigree**, referring to the adequacy, relevance, and coverage of the data used to develop and operate the M&S toolchain;
- **Traceability**, referring to the storage and documentation of datasets used throughout the toolchain lifecycle;
- **Uncertainty**, referring to the characterization of data acquisition methods, including known limitations of the data collection tools.

A second element addresses *personnel competency*, which is assessed at two complementary levels. At the organizational level, the framework considers the existence of established processes and procedures aimed at ensuring adequate M&S competencies within the organization. At the team level, the focus shifts to the specific skills and experience of the personnel responsible for developing, validating, and using the M&S toolchain for safety assessment purposes.

The third element relates to *release management* of the M&S toolchain. Given the dynamic nature of simulation environments and the frequency of toolchain updates, the framework requires management activities aligned with a structured product management approach. The scope of this activity is intentionally limited to M&S toolchain releases that directly contribute to the generation of certification-related evidence, excluding toolchain iterations used exclusively during ADS development or pre-assessment phases. This restriction aims to maintain a balanced approach that supports safety assurance without imposing unnecessary overhead. In fact, management activities are not limited to the technical content of the M&S toolchain itself, such as models and datasets, but also encompass the supporting software and hardware environment required for toolchain execution. These aspects are addressed through configuration management practices that ensure consistency, reproducibility, and transparency of simulation results.

### *3.2. M&S Analysis*

Complementing the management activities, the M&S analysis pillar addresses a central aspect of simulation credibility by focusing on the explicit description of the simulation toolchain and the justification of the modelling choices adopted for the intended use. As for the other pillars, its requirements are informed by the outcome of the criticality assessment. The objective of this pillar is to ensure transparency regarding how the M&S toolchain is constructed, what it is used for, and why it is appropriate for the associated level of decision criticality.

This phase also captures the use of multiple M&S toolchains within the same assessment context. In practice, manufacturers often rely on different virtual testing environments, each characterized by trade-offs between execution speed, cost, and modelling fidelity. Lower-fidelity tools are typically used to explore large scenario spaces and identify general performance trends, while higher-fidelity simulations are applied to a reduced set of scenarios to support statistically meaningful safety evaluations. In addition, M&S toolchains may adopt a modular structure, including so-called models of models, where multiple sub-models are combined to exploit the flexibility of virtual testing.

The *toolchain description* step provides a concise overview of the technical composition of the M&S toolchain and explains how the selected modelling approach aligns with the identified criticality level. This allows flexibility in the implementation of simulation solutions while requiring a clear and documented rationale for design choices.

A key element of the M&S analysis pillar is the systematic documentation of *assumptions and limitations* associated with the simulation models and toolchains. Making these aspects explicit ensures that known constraints are transparent to the personnel responsible for executing and interpreting virtual tests. Consistent with established M&S practices, modelling approaches can be broadly categorized as:

- **black-box**, typically lower-fidelity models reproducing input–output behaviour over a limited validity domain;
- **grey-box**, intermediate-fidelity models extending black-box representations with additional statistical or stochastic components;
- **white-box**, higher-fidelity models aimed at representing the underlying physical mechanisms to support robust extrapolation.

*3.3. M&S Verification*

The verification pillar addresses the numerical correctness and robustness of the models and algorithms implemented within the M&S toolchain. Its objective is to ensure that simulation outputs are not affected by implementation errors and that numerical results can be considered reliable inputs for subsequent validation and safety assessment activities. Conversely, the Validation pillar is in charge of verifying actual accuracy metrics, i.e., the discrepancy between virtual and real-world evidence. Fig 4 depicts a schematic representation of where the verification stands with respect to the overall V&V practices.

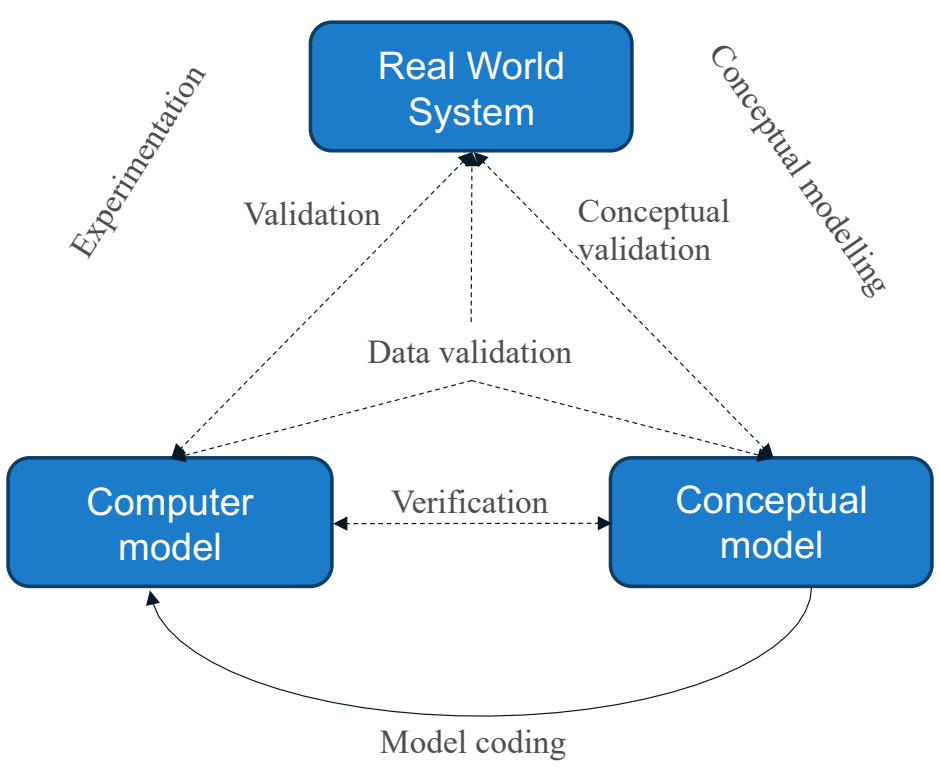

Fig 4. General V&V framework. Authors own adaptation based on (Oberkampf et al. 2004) .

The *code verification* phase focuses on confirming that the numerical algorithms underpinning the virtual models are correctly implemented and free from logical or numerical defects. This activity is typically performed by the software provider, as it is largely independent of a specific ADS application. Nevertheless, it remains the responsibility of the ADS certification applicant to obtain and document evidence that appropriate code verification procedures have been applied to the software components used within the M&S toolchain. Code verification comprises two complementary elements: numerical algorithm verification (NAV) and software quality assurance (SQA). NAV evaluates the correctness, accuracy, and numerical stability of the implemented algorithms, while SQA addresses software reliability aspects such as repeatability and consistency of results. Together, these activities establish a necessary foundation for the subsequent verification and validation steps.

The *calculation verification* phase assesses numerical errors arising from the specific realization of the simulation model, such as discretization in space or time, solver settings, or approximations of non-linear dynamics. Unlike code verification, this activity is model-specific and is therefore the responsibility of the simulation model developer. By explicitly quantifying numerical errors and solution accuracy, calculation verification ensures that the results produced by the M&S toolchain are sufficiently accurate for their intended use.

*Sensitivity analysis* (SA) evaluates the robustness of the M&S toolchain by assessing how variations in model parameters and operating conditions affect simulation outcomes. The objective is to identify parameters with the greatest influence on results and to understand the implications of their associated uncertainties. SA is performed within the validated domain of the M&S toolchain, avoiding parameter variations outside the range for which the simulation is considered applicable. Structured assessment approaches, such as multi-level scoring schemes proposed in established M&S credibility frameworks, can be used to support a transparent evaluation of robustness. Given the large number of parameters typically involved in ADS simulations, practical implementations focus on covering the most influential parameters rather than exhaustively analysing all possible factors.

### *3.4. M&S Validation*

The *validation pillar* of the M&S credibility framework addresses the extent to which simulation results are consistent with observations from the corresponding real-world system (RWS) (Dona and Ciuffo 2022). The objective is to determine whether the observed level of agreement is acceptable for the *intended use* of the M&S toolchain. Importantly, acceptance thresholds are not fixed a priori but emerge from the overall credibility assessment, taking into account the identified criticality and role of simulation within the safety evaluation process.

Validation activities may target individual subsystems or, where feasible, the integrated M&S toolchain. In addition, explicit attention is given to the validation domain, ensuring that any extrapolation beyond validated conditions is clearly identified and its associated risks understood. Agreement between simulation and RWS data can be quantified using a range of complementary approaches. In particular, statistical testing methods are well suited to the comparison of stochastic simulation environments (e.g., VeHiL) with stochastic real-world observations. The computation of the "agreement" can be carried out using several methods that are here summarized in Table 1 for the sake of convenience, together with examples.

Table 1. Validation comparison analyses methodologies summary.

| Comparison approach | Description | Example |
|---|---|---|
| *Graphical* | Qualitative evaluation of the signals | • Plotting M&S output vs. RWS output |
| *Scalar data* | Quantitative evaluation of scalar quantities deriving from either native scalar outputs or from time-series data following aggregation | • Comparison of minimum distance to an object at the end of a test (native scalar data)<br>• Aggregation of time-series using mean/median and other operators |
| *Time-series* | Quantitative evaluation of the discrepancies between two time-series using distance operators | • $L_2$ norm, Normalized Root Mean Square Error<br>• Sprague and Geers (Rabiner and Juang 1993) and Dynamic Time Warping<br>• Frequency domain approaches |
| *Statistical testing* | Quantitative verification of whether the null hypothesis "the model is an accurate representation of the real-world phenomena" cannot be rejected using statistical testing tools | • T-test or KS-test |

Because validation data are inherently limited, the fidelity of an M&S toolchain can only be established within a bounded parameter space. The framework therefore requires explicit documentation of the *validation domain* and its relationship to the intended application domain, including the degree of extrapolation involved.

The *subsystem validation* step focuses on validating individual components of the M&S toolchain. Its applicability depends on the overall toolchain architecture and the modelling approach adopted for each block. The final validation step addresses the performance of the *integrated M&S toolchain*. Depending on feasibility and toolchain structure, different validation strategies can be applied:

- **Replay with raw data**, comparing simulated sensor outputs against reconstructed real-world trajectories without ADS control;
- **Replay with perception**, comparing perception outputs in real and virtual environments along reconstructed trajectories;
- **Closed-loop simulation with ADS**, evaluating agreement at system level while the ADS actively controls the vehicle.

## 4. Conclusion

This paper has presented a risk-based credibility framework for qualifying M&S toolchains used in virtual testing of ADS. Motivated by the growing reliance on simulation to support safety-critical decisions, the framework adapts established practices from other safety-critical domains, most notably NASA STD-7009, to the specific challenges of automotive simulation and ADS assessment.

A central contribution of this work is the recognition that **simulation credibility is not a binary property**, but a context-dependent attribute linked to the *intended use* of simulation outputs and the *criticality* of the decisions they support. To address this, the framework structures credibility assessment around four complementary pillars—management, analysis, verification, and validation—providing a holistic alternative to traditional validation-only approaches. This structure enables proportional credibility requirements, supporting flexible yet rigorous use of virtual testing across different stages of safety assessment.

The framework has been developed through extensive interaction with stakeholders from industry, research, and public authorities, ensuring practical applicability while retaining methodological rigor. Its flexibility allows it to accommodate diverse simulation strategies, including multi-fidelity toolchains and modular "models of models", and to support both subsystem-level and integrated system validation. Although initially motivated by ADS applications, the approach is directly applicable to a broader range of road safety and simulation use cases, including advanced driver assistance systems (ADAS).

Despite these contributions, several limitations remain. The effectiveness of the framework depends on the availability of suitable validation data, transparent documentation of modelling assumptions, and sufficient expertise to perform criticality, uncertainty, and sensitivity analyses. Moreover, the framework has not yet been systematically benchmarked across a large set of independent simulation toolchains or modelling paradigms, particularly those relying heavily on data-driven or learning-based approaches. As virtual testing continues to expand its role in road safety assessment, structured credibility frameworks such as the one presented here will be essential to ensure that simulation-based evidence remains transparent, trustworthy, and fit for purpose.

## Acknowledgements

This research has been funded by the Joint Research Centre for the European Commission, Ispra (VA), Italy, and by the Directorate General for Internal Market, Industry, Entrepreneurship and SMEs for the European Commission, Brussels, Belgium. The opinions expressed in the manuscript are those of the authors and may not in any circumstances be considered to represent an official opinion of the European Commission.